\documentclass[letterpaper]{article} 
\usepackage{aaai25}  
\usepackage{times}  
\usepackage{helvet}  
\usepackage{courier}  
\usepackage[hyphens]{url}  
\usepackage{graphicx} 
\usepackage{natbib}  
\usepackage{caption} 
\usepackage{algorithm}
\usepackage{algorithmic}
\usepackage{multirow}
\usepackage{colortbl}
\usepackage{subcaption}
\usepackage{array}
\usepackage{amsmath}
\usepackage{xcolor}
\usepackage{booktabs}
\usepackage{amssymb}
\usepackage[T1]{fontenc}

\usepackage{newfloat}
\usepackage{listings}
\DeclareCaptionStyle{ruled}{labelfont=normalfont,labelsep=colon,strut=off} 
\floatstyle{ruled}
\newfloat{listing}{tb}{lst}{}
\floatname{listing}{Listing}
\title{EchoIR: Advancing Image Restoration with Echo Upsampling and Bi-Level Optimization}
\author{
    Yuhan He\textsuperscript{\rm 1}, Yuchun He\textsuperscript{\rm 1}
}
\affiliations{
    \textsuperscript{\rm 1}University of Sydney\\

    Camperdown NSW 2050, Australia\\
    \{yuhe0760, yuhe0953\}@uni.sydney.edu.au
}

\usepackage{bibentry}

\begin{document}

\maketitle

\begin{abstract}
Image restoration represents a fundamental challenge in low-level vision, focusing on reconstructing high-quality images from their degraded counterparts. With the rapid advancement of deep learning technologies, transformer-based methods with pyramid structures have advanced the field by capturing long-range cross-scale spatial interaction. Despite its popularity, the degradation of essential features during the upsampling process notably compromised the restoration performance, resulting in suboptimal reconstruction outcomes. We introduce the EchoIR, an UNet-like image restoration network with a bilateral learnable upsampling mechanism to bridge this gap. Specifically, we proposed the Echo-Upsampler that optimizes the upsampling process by learning from the bilateral intermediate features of U-Net, the "Echo", aiming for a more refined restoration by minimizing the degradation during upsampling. In pursuit of modeling a hierarchical model of image restoration and upsampling tasks, we propose the Approximated Sequential Bi-level Optimization (AS-BLO), an advanced bi-level optimization model establishing a relationship between upsampling learning and image restoration tasks. 
Extensive experiments against the state-of-the-art (SOTA) methods demonstrate the proposed EchoIR surpasses the existing methods, achieving SOTA performance in image restoration tasks. 
\end{abstract}

%
\begin{figure}[t]
\scriptsize
\centering
\includegraphics[width=0.9\linewidth]{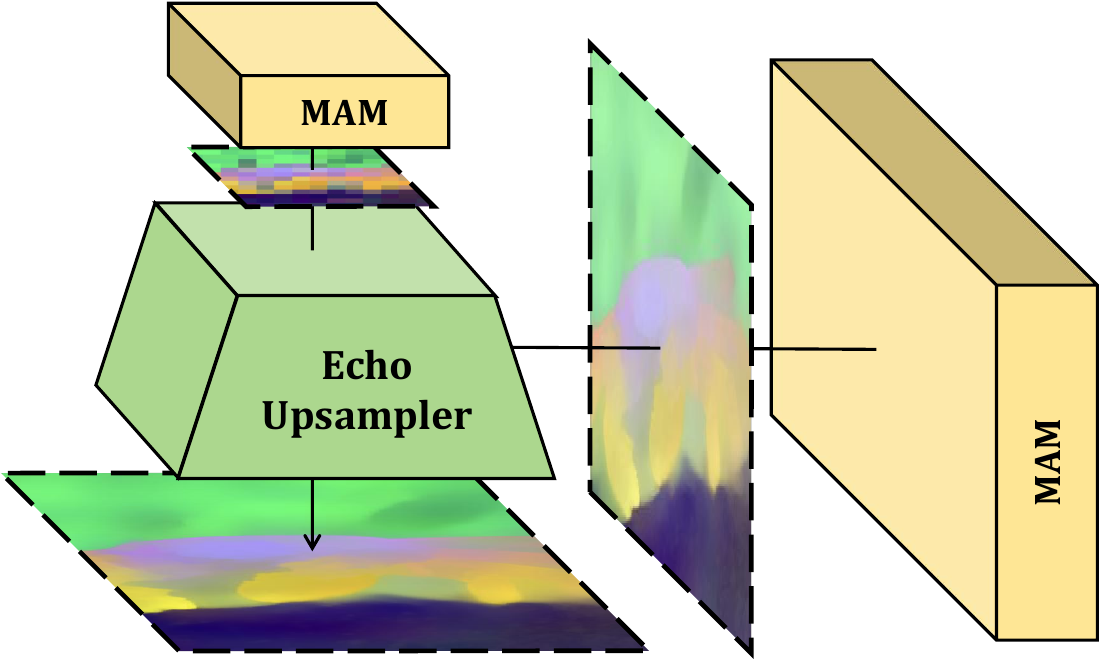}\\
(a) Echo-Upsampler\\
\vspace{0.15in}
\begin{tabular}{cc}
\includegraphics[width=0.20\textwidth]{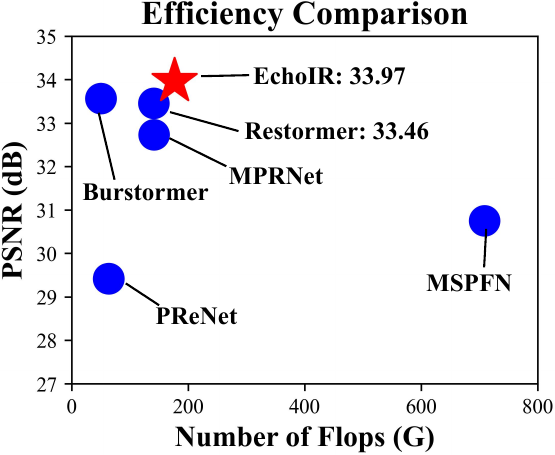} &
\includegraphics[width=0.20\textwidth]{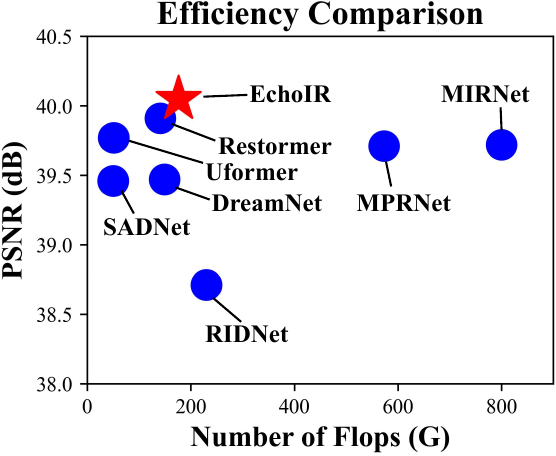}
\\
(b) Deraining (Table \ref{table:derain_metrics}) &
(c) Denoising (Table \ref{table:denoise_metrics}) 
\end{tabular}
\caption{Schematic diagram of the Echo-Upsampler and comparison of the EchoIR model with other methods in deraining and denoising tasks. MAM is Mix-attention module.}
\label{fig:head}
\end{figure}

\section{Introduction}

Image restoration stands as a fundamental image-processing task to reconstruct high-quality images from their degraded counterparts \cite{li2025contourlet, li2023text, jiang2024towards, zou2024enhancing, zou2024contourlet}. However, the challenge of image restoration primarily stems from the loss or absence of information inherent in the input image, such as noise and blur, which can obscure essential image details. Effectively distinguishing between image degradation and the main information of the image, and subsequently restoring the lost information, has become a formidable problem. With the rise of deep learning, researchers began exploring its applications in image-related tasks. Convolutional Neural Networks (CNN) \cite{CNN} introduced new methods for image restoration \cite{DreamNet, DMPHN}. However, CNNs have limited capability in capturing long-range dependencies on pixel information, restricting their effectiveness. Transformers establish feature dependencies through attention mechanisms, leading to promising results in visual tasks \cite{SwinIR, MPRNet}. Despite this, Transformers suffer from quadratic growth in computational complexity with increasing spatial resolution. To improve efficiency, solutions such as spatial window attention and combining multiple attention mechanisms have been proposed \cite{ELAN, DAT}.

Existing image restoration methods often face challenges with information loss during downsampling, leading to inadequate upsampling and poor restoration outcomes. Downsampling reduces feature map size, losing crucial information and impacting recovery. Although U-Net \cite{U-Net, restormer, MPRNet} and ResNet \cite{ResNet, SwinIR, ELAN, DAT} structures help retain features, the retained information is often insufficient.

To address the aforementioned challenges, we propose EchoIR, a bilaterally learnable upsampling image restoration network built upon the U-Net architecture. We refer to the feature information obtained during the U-Net downsampling process as "echo", which is subsequently utilized during the upsampling phase. Specifically, we introduce the Echo-Upsampler, which learns the upsampling process by leveraging the "echo" derived from U-Net downsampling encoding, thereby enhancing the quality of upsampling.

To enhance the image restoration and upsampling tasks, we introduce a hyperparameter optimization strategy to simplify the bi-layer optimization model. This strategy breaks down the complex Echo-Upsampler and image restoration optimization into single-layer problems solvable via gradient descent. Our proposed Approximated Sequential Bi-level Optimization (AS-BLO) reformulates the complex bi-level model into non-constraint single-level tasks, which are easier to solve by gradient descent, and proves the asymptotic convergence of these approximated problems.

By employing Echo-Upsampler and AS-BLO, we efficiently preserve nuanced features, thereby enhancing the quality of upsampling and image restoration. Additionally, our proposed hyperparameter optimization strategy for reformulating the bi-level optimization model ensures the execution of an optimal upsampling process and minimizes image recovery loss. Our contributions are as follows:
\begin{itemize}
\item To address the issue of image information loss during the upsampling process in existing methods, we introduce the Echo-Upsampler. This component optimizes the upsampling procedure by learning from the bilateral intermediate features of U-Net.
\item We proposed the Approximated Sequential Bi-level Optimization (AS-BLO) to model the problems of image restoration and Echo-Upsampler learning. By transforming it into a sequential single-layer optimization problem, we solve the original complex hierarchical model using gradient descent.
\item Extensive experiments prove that the proposed Echo-Net achieves state-of-the-art (SOTA) recovery results in the image denoising, deraining, and deblurring tasks.
\end{itemize}

\section{Related Work}

\subsection{Image Restoration}
The field of image restoration has experienced a transformation with the integration of deep learning technologies. Denoising, deblurring, and deraining are three essential tasks in the field of image restoration. Denoising involves the removal of noise from images. Noise can occur due to various factors, including sensor imperfections, environmental conditions during the image capture process, or data transmission errors. The objective of denoising models\cite{dncnn, restormer} is to recover the original, clean image by eliminating this noise while preserving important details and structures. Deblurring is the process of removing blur from images. Blur can be caused by several factors, such as camera or subject movement during exposure, out-of-focus shooting, or atmospheric disturbances. Deblurring models \cite{Uformer,  stripformer} aim to sharpen the image details that have been smeared or obscured due to blur, restoring clarity and improving the overall image quality. Deraining removes rain streaks or droplets that degrade image quality. Deraining methods \cite{IPT, restormer} enhance visual appeal and usability by eliminating these artifacts. Among the trailblazing models that have set the foundation for these advancements, SRCNN \cite{SRCNN} and ARCNN \cite{ARCNN} stand out as pioneering efforts in applying convolutional neural networks (CNNs) for image restoration tasks.

\begin{figure*}[t]
    \centering
    \includegraphics[width=0.95\linewidth]{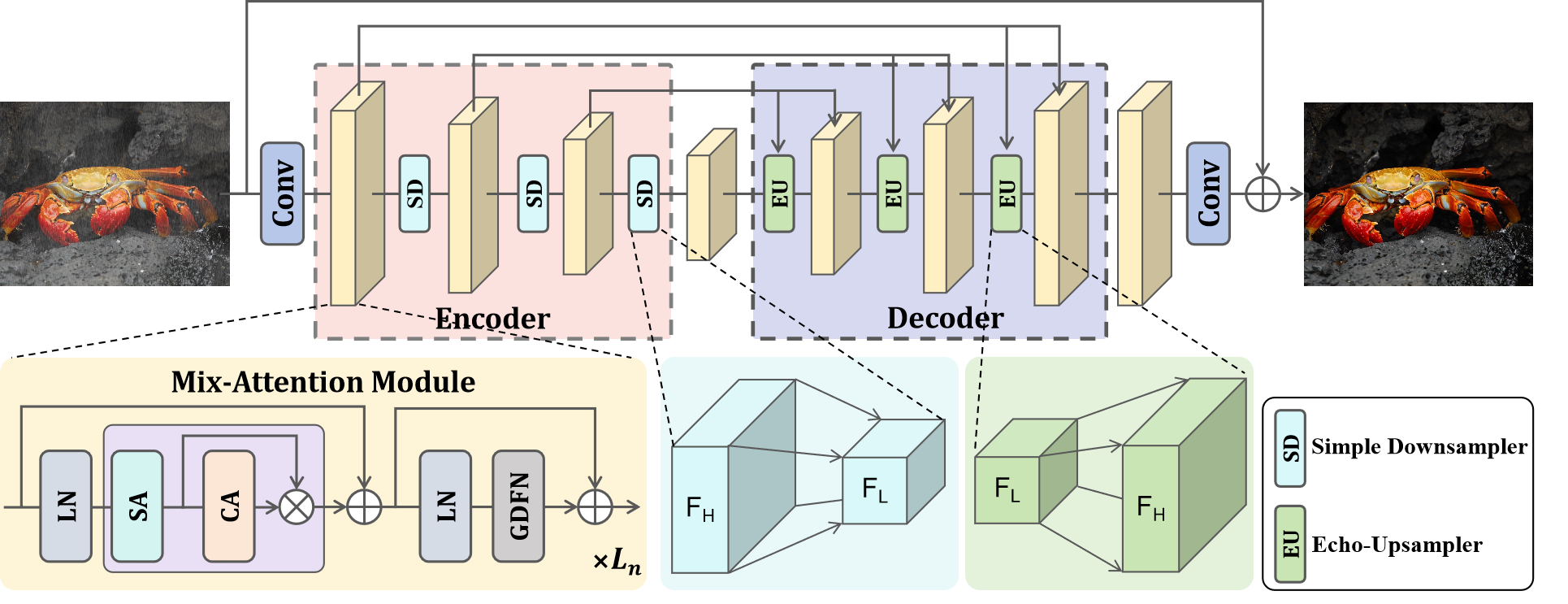}
    \caption{The structure of EchoIR. SD is the simple downsampler. EU is Echo-upsampler. LN is layer normalization, SA is multi-head self-attention, CA is channel attention, and GDFN is gated-Dconv FN.}
    \label{EchoIR}
\end{figure*}



\subsection{Transformer-based Restoration}

These initial models laid the groundwork for subsequent advancements by incorporating sophisticated CNN architectures and techniques like residual connections \cite{CASCNN, Kim_2016_CVPR, DePrior}. However, CNN-based methods often struggle with modeling global dependencies effectively. Generative adversarial networks (GANs) \cite{NIPS2017_892c3b1c, Wang_2018_ECCV_Workshops} have also been used for image restoration but face challenges such as mode collapse, difficult training processes, and potential inaccuracies, leading to ongoing research for improvements.

These challenges have spurred the exploration of Transformers as a powerful alternative due to their ability to capture global dependencies effectively. Initially applied in natural language processing, Transformers \cite{Transformer} have proven effective in computer vision through models like Vision Transformer (ViT) \cite{ViT}. Despite their strengths, Transformers in image restoration, such as IPT \cite{IPT} and SwinIR \cite{SwinIR}, face challenges with quadratic computational complexity in their self-attention mechanisms. IPT tackles this by processing smaller image patches independently, while SwinIR uses shifted window attention for enhanced performance. Further advancements \cite{chen2023activating, chen2024recursive, chen2023dual, li2022efficient, restormer} have been made in developing effective attention mechanisms for restoration. However, current methods have not effectively addressed the issue of information loss during upsampling.

\section{Method}
The architectural framework of EchoIR is delineated in Figure \ref{EchoIR}, comprising four principal components: shallow feature extraction, feature encoder, feature decoder, and image reconstruction. Shallow feature extraction entails acquiring multi-channel feature maps through convolutional operations. The feature encoder integrates mix-attention modules and downsamplers to facilitate deep feature extraction across different levels. Each mix-attention module encompasses multiple mix-attention blocks to enable deep feature extraction at every level. The feature decoder employs Mix-Attention modules and Echo-Upsamplers to execute feature extraction and restore the feature map to the dimensions of shallow features. The Echo-Upsampler expands the current features guided by the encoder layer features. Leveraging the AS-BLO optimization method enhances the training of the upper echo sampler. Ultimately, image reconstruction transpires to convert feature maps into restored images.

\subsection{Overall Architecture}
Given a low-quality image $I_{L}\in\mathbb{R}^{3\times H\times W}$ ($H$ is the height of the image, $W$ is the width of the image), EchoIR first generates a feature map $F\in\mathbb{R}^{C\times H\times W}$ by shallow feature extraction. The feature encoder processes the feature map by a series of mix-attention modules (MAM) with length $[L_1,\ L_2,\ ...,\ L_n]$. The shape of the feature map before and after each MAM remains consistent. After a MAM processes the feature map, it goes through a simple downsampler, and its height and width will be half of the original, and the number of channels will be doubled. After passing through the feature encoder, the feature map becomes $F_e\in\mathbb{R}^{2C\times \frac{H}{8}\times \frac{W}{8}}$. After a middle block, the feature map is passed to the decoder. The feature decoder also uses the same numbers of MAMs as the encoder to process the feature map. The lengths of the models are $[L_{n},\ L_{n-1},\ ...,\ L_1]$. Except for the last MAM, the feature map passes through an Echo-Upsampler before entering the MAM. The Echo-Upsampler will double the height and width of the feature map and reduce the number of channels to half. The feature map after the feature decoder is $Fd\in\mathbb{R}^{C\times H\times W}$ 

\subsection{Mix-Attention Block}
Compared with the general attention structure, mix-attention combines channel attention and multi-head self-attention. The feature map $F_t$ is first subjected to layer normalization when entering the Mix-attention Block. During the multi-head self-attention process, $F_t$ undergoes a convolution operation to triple the number of channels. Divide the $F_t$ into three parts, Q, K, and V, and divide these three parts into multiple heads. Do the following for Q, K, and V:
\begin{equation}
\begin{gathered}
Alpha(Q,\ K)=\frac{(Q\cdot K)}{\sqrt{d}},\\
\\
Attention(Q,\ K,\ V)=V\cdot Softmax(Alpha(Q,\ K)).
\end{gathered}
\label{attention}
\end{equation}
After multi-head self-attention, the Mix-attention Block gets $F_{ts}$ and enters the channel attention process.

$F_ts$ first obtains a one-dimensional feature $pw\in \mathbb{R}^{C}$ through adaptive average pooling. $pw$ gets $cw\in \mathbb{R}^{C}$ after passing through a multi-layer perceptron and Sigmoid heuristic function. Perform element-level multiplication of feature maps $F_ts$ and $cw$ to obtain $F_{tc}$. Perform element-level addition of $F_{tc}$ and $F_t$ to obtain $F_{ta}$. The process is shown as:
\begin{equation}
\begin{gathered}
pw=AdaptiveAvgPool(F_{ts}),\\
cw=Sigmoid(Mlp(pw)),\\
F_{tc}=F_{ts}\otimes cw,\ F_{ta}=F_{tc}\oplus F_{t}.
\label{channel attention}
\end{gathered} 
\end{equation}

Then $F_{ta}$ undergoes layer normalization and gated-Dconv FN (GDFN) \cite{restormer} to obtain $F_{tg}$. GDFN can be used to filter effective information for further feature extraction. Perform element-level addition of $F_{tg}$ and $F_{ta}$ to obtain $F_{t+1}$. $F_{t+1}$ will enter the next Mix-attention Block or other modules.
\begin{equation}
\begin{gathered}
pw=AdaptiveAvgPool(F_{ts}),\\
cw=Sigmoid(Mlp(pw)),\\
F_{tc}=F_{ts}\otimes cw.
\label{GDFN}
\end{gathered}
\end{equation}

\subsection{Echo-Upsampler}
During the upsampling process, the Echo-Upsampler uses the output features of the Mix-attention module at the same level in the encoder as a reference to perform the upsampling operation. It is similar to joint bilateral upsampling (JBU) \cite{JBU}. JBU can produce smoother and higher-resolution results with fewer parameters. The reference input in Echo-Unpsampler is a feature map but not a high-quality image. The workflow of the Echo-Upsampler is shown in figure \ref{fig:head}. The process is as follows:
\begin{equation}
\begin{gathered}
W(p,p') = f(p,p')\oplus g(F_{ref}[p],F_{ref}[p']),\\
F_{up}[p]=\frac{1}{sum(W(p,p'))}\sum_{p'\in\Omega}F_{down}[p']\cdot W(p,p').
\label{EchoUp}
\end{gathered}
\end{equation}
$F_{down}$ is the feature map before upsample, and $F_{up}$ is after upsample. $F_{ref}$ is the referenced feature map from the Mix-Attention Modules in the encoder. Different Echo-upsamplers use the feature map of the corresponding level of the encoder as a reference. $\Omega$ is a set of pixels within a certain range centered on pixel $p$. $f$ and $g$ represent two filter kernels: spatial kernel and range kernel. $f$ is a Gaussian kernel which is shown as:
\begin{equation}
\begin{gathered}
f(p,p')=e^{-\frac{\Vert p-p'\Vert^2}{2\sigma^2}}.
\label{Gaussian}
\end{gathered}
\end{equation}
where parameter $\sigma$ is learnable. $g$ is also calculated using the Gaussian kernel function. Since $F_{ref}$ is not a high-definition lossless image, a multi-layer perceptron is added on this basis to adjust $F_{ref}$. The process is shown as:
\begin{equation}
\begin{gathered}
g(F_{ref}[p],F_{ref}[p'])=e^{-\frac{\Vert Mlp(F_{ref}[p])-Mlp(F_{ref}[p'])\Vert^2}{2\sigma^2}}.
\label{Gaussian}
\end{gathered}
\end{equation}

Through this sampling method, a higher-quality output can be obtained when the input features are not smooth enough. This is beneficial to the subsequent Mix-Attention module for feature extraction.

\subsection{Approximated Sequential Bi-Level Optimization}
Our network is architectured to optimize image reconstruction as its upper-level (UL) goal while taking image restoration and a lossless Echo-Upsampler as foundational lower-level (LL) objectives. This configuration forms a hierarchical bi-level optimization framework, where the UL objective synergizes with the LL solutions to enhance the overall reconstruction quality:

\begin{equation}
\begin{gathered}
\min _{\mathbf{\beta} \in \mathcal{B}} \min _{\mathbf{\omega} \in \mathbb{R}^n} F(\mathbf{\beta}, \mathbf{\omega}; \mathcal{D}_{val}), \\
\text {s.t. } \mathbf{\omega} \in \mathcal{S}(\mathbf{\beta}) := \arg \min_{\omega} f(\mathbf{\beta}, \mathbf{\omega}; \mathcal{D}_{tr}),
\end{gathered}
\label{general BLO}
\end{equation}

\noindent where \(\beta\) is the UL hyperparameter, \(\omega\) is the LL parameter. \(\mathcal{S}(\mathbf{\beta})\) is the solution set of the LL objective. The notations \(F(\cdot)\) and \(f(\cdot)\) correspond to the objectives of the UL and LL, respectively. The datasets \(\mathcal{D}_{val}\) and \(\mathcal{D}_{tr}\) delineate the distinct sets of data reserved for validation and training purposes, respectively. To solve the LL convexity assumption and the possible non-singleton \(\mathcal{S}(\mathbf{\beta})\), we transform the complex bi-level optimization model in Eq.~\ref{general BLO} into a concise version:

\begin{equation}
\begin{gathered}
\min _{\mathbf{\beta} \in \mathcal{B}} \varphi(\mathbf{\beta; \mathcal{D}_{val}}),\\
\quad \varphi(\mathbf{\beta; \mathcal{D}_{val}}):=\min _{\mathbf{\omega}}\{F(\mathbf{\beta}, \mathbf{\omega}; \mathcal{D}_{val}): \mathbf{\omega} \in \mathcal{S}(\mathbf{\beta})\},
\end{gathered}
\label{simple BLO}
\end{equation}

\noindent where \(\varphi(\cdot)\) is the value-function of the LL objective. For a fixed \(\beta\), solving \(\varphi(\cdot)\) is a tractable bi-level optimization problem with one variable \(\omega\). We simplify it further to a series of single-level problems using the value function of the LL problem:

\begin{equation}
f^{*}(\mathbf{\beta}; \mathcal{D}_{tr}) :=  \min_{\omega} f(\mathbf{\beta}, \mathbf{\omega}; \mathcal{D}_{tr}).
\label{value function1}
\end{equation}

Then the model can be further reformulated as:

\begin{equation}
\begin{gathered}
\varphi(\mathbf{\beta; \mathcal{D}_{val}})=\\
\min _{\mathbf{\mathbf{\omega}}}\left\{F(\mathbf{\beta}, \mathbf{\omega}; \mathcal{D}_{val}) : f(\mathbf{\beta}, \mathbf{\omega};\mathcal{D}_{tr}) \leq  f^{*}(\mathbf{\beta}; \mathcal{D}_{tr})\right\}.
\label{inequality constraint}
\end{gathered}
\end{equation}

We then add a barrier function and a regularization term \(\frac{\mu}{2}\|\mathbf{\omega}\|^2, \mu > 0 \), reformulating the model as follows:

\begin{equation}
\begin{gathered}
\varphi_{\mu, \theta, \sigma}(\mathbf{\beta; \mathcal{D}_{val}})=\\
\min _{\mathbf{\omega}}\left\{F(\mathbf{\beta}, \mathbf{\omega}; \mathcal{D}_{val})+P_\sigma\left(\mathbf{\beta, \omega; \mathcal{D}_{tr}}\right)+\frac{\theta}{2}\|\mathbf{\omega}\|^2\right\},
\end{gathered}
\label{final simplification}
\end{equation}

\begin{equation}
P(\zeta; \sigma)=\left\{\begin{array}{lr}
-\sigma\left(\log (-\zeta)+\eta_1\right), & -\kappa \leq \zeta<0 \\
-\sigma\left(\eta_2+\frac{\eta_3}{\zeta^2}+\frac{\eta_4}{\zeta}\right), & \zeta<-\kappa \\
\infty, & \zeta \geq 0
\end{array}\right.
\label{barrier function}
\end{equation}

\noindent where in Eq.~\ref{final simplification}, the \(P_\sigma\left(\mathbf{\beta, \omega; \mathcal{D}_{tr}}\right)\) denoted as \(P_\sigma(f(\mathbf{\beta, \omega; \mathcal{D}_{tr}})-\) \(f_\mu^*(\mathbf{\beta; \mathcal{D}_{tr}}))\) is the barrier function with parameter \(\sigma\) for sequential unconstrained minimization. \(0 < \kappa \leq 1, \eta_1, \eta_2, \eta_3, \eta_4\) are chosen such that \(P_{\sigma} \geq 0\) and is twice differentiable. \(\frac{\theta}{2}\|\mathbf{\omega}\|^2\) stands for the regularization term and \((\mu, \theta, \sigma) > 0\). This reformulates the constraint problem in Eq.~\ref{inequality constraint} into a sequence of unconstrained converge single-level problems.


Recall that the \(\omega \in S(\beta)\) is unique, denote as \(\omega^{*}(\beta)\), then the approximated \(\nabla_{\mathbf{\beta}} \varphi(\mathbf{\beta; \mathcal{D}_{val}})\) is calculated by chain rule:

\begin{equation}
\begin{gathered}
\nabla_{\mathbf{\beta}} \varphi(\mathbf{\beta; \mathcal{D}_{val}}) = \nabla_{\beta} F(\mathbf{\beta}, \mathbf{\omega^{*}}; \mathcal{D}_{val}) + G(\beta),\\
\text {s.t. } G(\beta) = (\nabla_{\beta} \omega^{*})^{T} \cdot \nabla_{\omega} F(\mathbf{\beta}, \mathbf{\omega^{*}}; \mathcal{D}_{val}),
\end{gathered}
\label{d_phi}
\end{equation}

\noindent where the \(\nabla_{\beta} F(\mathbf{\beta}, \mathbf{\omega^{*}}; \mathcal{D}_{val})\) is explicit gradient, \(G(\beta)\) is implicit gradient, so only \(G(\beta)\) is what we after to solve. We denote:

\begin{equation}
\mathbf{z}_\mu^*(\mathbf{\beta})=\underset{\mathbf{\omega}}{\operatorname{argmin}}\left\{f(\mathbf{\beta, \omega; \mathcal{D}_{tr}})+\frac{\mu}{2}\|\mathbf{\omega}\|^2\right\},
\label{z*}
\end{equation}

\begin{equation}
\begin{gathered}
\omega^{*}_{\mu, \theta, \sigma}(\mathbf{\beta})=\\
\underset{\mathbf{\omega}}{\operatorname{argmin}}\left\{F(\mathbf{\beta}, \mathbf{\omega}; \mathcal{D}_{val})+P_\sigma\left(\mathbf{\beta, \omega; \mathcal{D}_{tr}}\right)+\frac{\theta}{2}\|\mathbf{\omega}\|^2\right\},
\end{gathered}
\label{final simplification w}
\end{equation}

\noindent where the \(P_\sigma\left(\cdot\right)\) in Eq.~\ref{final simplification w} stands for \(P_\sigma(f(\mathbf{\beta, \omega; \mathcal{D}_{tr}})-\) \(f_\mu^*(\mathbf{\beta; \mathcal{D}_{tr}}))\). Given \(\beta \in \mathcal{B}\), and \(\mu, \theta, \sigma > 0\), \(\mathbf{z}_\mu^*(\mathbf{\beta})\) and \(\omega^{*}_{\mu, \theta, \sigma}(\mathbf{\beta})\) are singleton, we then have \(\varphi_{\mu, \theta, \sigma}(\mathbf{\beta; \mathcal{D}_{val}})\) differentiable and \(G(\beta)\) can be solved:

\begin{equation}
G(\mathbf{x})=\nabla_{\mathbf{\beta}} P_\sigma(f(\mathbf{\beta, \omega; \mathcal{D}_{tr}})-f_\mu^*(\mathbf{\beta; \mathcal{D}_{tr}}))|_{\mathbf{\omega}=\mathbf{\omega}_{\mu, \theta, \sigma}^*(\mathbf{\beta})},
\label{solution}
\end{equation}

\noindent where \(f_\mu^*(\mathbf{\beta; \mathcal{D}_{tr}}) = f\left(\mathbf{\beta}, \mathbf{z}_\mu^*(\mathbf{\beta}); \mathcal{D}_{tr}\right)+\frac{\mu}{2}\left\|\mathbf{z}_\mu^*(\mathbf{\beta})\right\|^2\) and \(\nabla_{\beta}f_\mu^*(\mathbf{\beta; \mathcal{D}_{tr}}) = \nabla_{\beta}f_\mu^*(\mathbf{\beta, \omega; \mathcal{D}_{tr}})|_{\omega = \mathbf{z}_\mu^*(\mathbf{\beta}) }\). Since we have \(f(\beta, \omega;\mathcal{D}_{tr})\) is level-bounded in \(\omega\) locally uniformly in \(\beta \in \mathcal{B}\), the inf-compactness condition holds for \(f\left(\cdot\right) + \frac{\mu}{2}\|\mathbf{\omega}\|^2\). 

\section{Experiments}

\begin{table*}[!t]
\centering
  \footnotesize
  \renewcommand\arraystretch{1.1} 
	\setlength{\tabcolsep}{0.3mm}
  \caption{Comparison with the deraining methods. The best result is in \textcolor{red}{red} and the second one is in \textcolor{blue}{blue}.}
  \vspace{-0.1in}
  \resizebox{\textwidth}{!}{ 
    \begin{tabular}{l|>{\centering\arraybackslash}m{1.1cm}>{\centering\arraybackslash}m{1.1cm}>{\centering\arraybackslash}m{1.1cm}>{\centering\arraybackslash}m{1.1cm}>{\centering\arraybackslash}m{1.1cm}>{\centering\arraybackslash}m{1.1cm}>{\centering\arraybackslash}m{1.1cm}>{\centering\arraybackslash}m{1.1cm}>{\centering\arraybackslash}m{1.1cm}>{\centering\arraybackslash}m{1.1cm}||>{\centering\arraybackslash}m{1.1cm}>{\centering\arraybackslash}m{1.1cm}}
    \toprule
     \cellcolor{gray!10} &  \multicolumn{2}{c}{\cellcolor{gray!10}Test100} & \multicolumn{2}{c}{\cellcolor{gray!10}Rain100H } & \multicolumn{2}{c}{\cellcolor{gray!10}Rain100L } & \multicolumn{2}{c}{\cellcolor{gray!10}Test2800} & \multicolumn{2}{c||}{\cellcolor{gray!10}Test1200 } & \multicolumn{2}{c}{\cellcolor{gray!10}Average} \\
        \cellcolor{gray!10}\multirow{-2}{*}{Methods} & \cellcolor{gray!10}PSNR$\uparrow$ & \cellcolor{gray!10}SSIM $\uparrow$ & \cellcolor{gray!10}PSNR $\uparrow$ & \cellcolor{gray!10}SSIM $\uparrow$ & \cellcolor{gray!10}PSNR $\uparrow$ & \cellcolor{gray!10}SSIM $\uparrow$ & \cellcolor{gray!10}PSNR $\uparrow$ & \cellcolor{gray!10}SSIM $\uparrow$ & \cellcolor{gray!10}PSNR $\uparrow$ & \cellcolor{gray!10}SSIM $\uparrow$ & \cellcolor{gray!10}PSNR $\uparrow$ & \cellcolor{gray!10}SSIM $\uparrow$\\
\midrule
    DerainNet \cite{DerainNet} & 22.77  & 0.810  & 14.92  & 0.592  & 27.03  & 0.884  & 24.31  & 0.861  & 23.38  & 0.835  & 22.48  & 0.796  \\
    SEMI \cite{semi} & 22.35  & 0.788  & 16.56  & 0.486  & 25.03  & 0.842  & 24.43  & 0.782  & 26.05  & 0.822  & 22.88  & 0.744  \\
    PReNet \cite{PReNet} & 24.81  & 0.851  & 26.77  & 0.858  & 32.44  & 0.950  & 31.75  & 0.916  & 31.36  & 0.911  & 29.42  & 0.897  \\
    MSPFN \cite{MSPFN} & 27.50  & 0.876  & 28.66  & 0.860  & 32.40  & 0.933  & 32.82  & 0.930  & 32.39  & 0.916  & 30.75  & 0.903  \\
    MPRNet \cite{MPRNet} & 30.27  & 0.897  & 30.41  & 0.890  & 36.40  & 0.965  & 33.64  & 0.938  & 32.91  & 0.916  & 32.73  & 0.921  \\
    MFDNet \cite{MFDNet} & 30.78  & 0.914  & 30.48  & 0.899  & \textcolor{blue}{37.61}  & 0.973  & 33.55  & 0.939  & 33.01  & 0.925  & 33.08  & 0.930  \\
    DGUNet \cite{DGUNet} & 30.32  & 0.899  & 30.66  & 0.891  & 37.42  & 0.969  & 33.68  & 0.938  & 33.23  & 0.920  & 33.06  & 0.923  \\
    SPAIR \cite{SPAIR} & 30.35  & 0.909  & 30.95  & 0.892  & 36.93  & 0.969  & 33.34  & 0.936  & 33.04  & 0.922  & 32.91  & 0.926  \\
    Restormer \cite{restormer} & 31.63  & 0.923  & 31.14  & 0.907  & 37.46  & 0.972  & 33.98  & 0.942  & 33.11  & 0.926  & 33.46  & 0.934  \\
    Burstormer \cite{Burstormer} & \textcolor{blue}{31.72}  & \textcolor{blue}{0.930}  & \textcolor{blue}{31.20}  & \textcolor{blue}{0.911}  & 37.59  & \textcolor{blue}{0.982}  & \textcolor{blue}{34.09}  & \textcolor{blue}{0.949}  & \textcolor{blue}{33.25}  & \textcolor{blue}{0.932}  & \textcolor{blue}{33.57}  & \textcolor{blue}{0.941}  \\
    \midrule
    \textbf{EchoIR (ours)}  & \textcolor{red}{32.17}  & \textcolor{red}{0.939}  & \textcolor{red}{31.58}  & \textcolor{red}{0.926}  & \textcolor{red}{38.04}  & \textcolor{red}{0.993}  & \textcolor{red}{34.41}  & \textcolor{red}{0.953}  & \textcolor{red}{33.66}  & \textcolor{red}{0.947}  & \textcolor{red}{33.97}  & \textcolor{red}{0.952}  \\
    \bottomrule
    \end{tabular}%
    }
  \label{table:derain_metrics}%
\end{table*}

\begin{figure*}[!h]
\footnotesize
\centering
\renewcommand\arraystretch{1.5} 
\scalebox{0.92}{
\begin{tabular}{cccc}
\hspace{-0.42cm}
\begin{tabular}{c}
\includegraphics[width=0.32\textwidth]{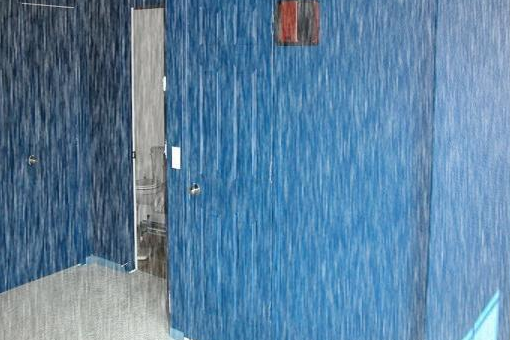}
\\
Raining Image
\end{tabular}
\hspace{-0.1cm}
\begin{tabular}{cccccc}
\includegraphics[width=0.13\textwidth]{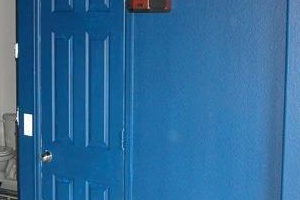} \hspace{-1.5mm} &
\includegraphics[width=0.13\textwidth]{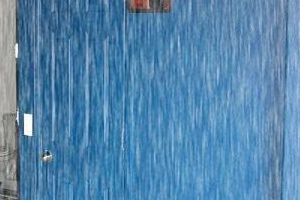} \hspace{-1.5mm} &
\includegraphics[width=0.13\textwidth]{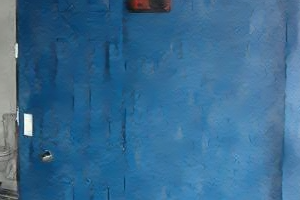} \hspace{-1.5mm} &
\includegraphics[width=0.13\textwidth]{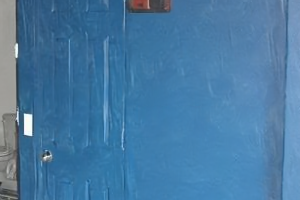} \hspace{-1.5mm}
\\
Gt \hspace{-1.5mm} &
Raining \hspace{-1.5mm} &
PReNet \hspace{-1.5mm} &
MPRNet \hspace{-1.5mm}
\\
\includegraphics[width=0.13\textwidth]{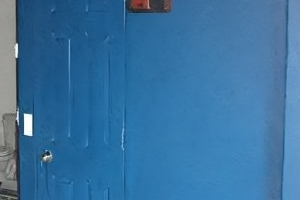} \hspace{-1.5mm} &
\includegraphics[width=0.13\textwidth]{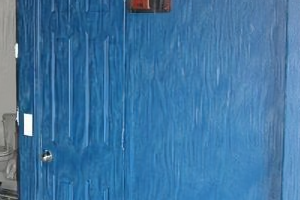} \hspace{-1.5mm} &
\includegraphics[width=0.13\textwidth]{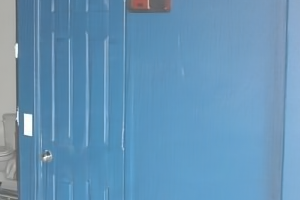} \hspace{-1.5mm} &
\includegraphics[width=0.13\textwidth]{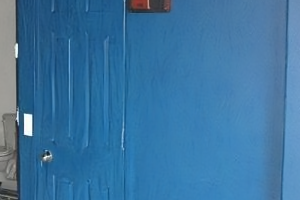} \hspace{-1.5mm} 
\\ 
MFDNet \hspace{-1.5mm} &
DGUNet \hspace{-1.5mm} &
Restormer \hspace{-1.5mm} &
Ours \hspace{-1.5mm} 
\\
\end{tabular}
\\
\end{tabular} }
\vspace{-0.1in}
\caption{Visual comparison of results with SOTA draining methods. This comparison employs the output results for direct comparison, thereby providing a clear reflection of the quality of the model's output outcomes.}
\label{fig:derainning_visual}
\end{figure*}

\subsection{Experiments Setting}
For the structure of EchoIR in this experiment, the lengths of the Mix-Attention modules in the encoder are 4, 6, and 6 respectively, and the feature map dimensions for each module are 48, 96, and 192. The structure of the decoder is the same as that of the encoder but in reverse order. The length of the module in the middle block is 8, and the dimension of the feature map is 384. The final processing module length is 4, and the feature map dimension is 48. The expansion factor of GDFN is 2.66. In the experiment, the number of iterations is 3e+5. The stages of iterations are divided into [92000,64000,48000,36000,36000,24000], the corresponding patch size is [144, 160, 192, 224, 320, 336] and the batch size is [6, 5, 4, 2, 1, 1]. The initial learning rate of training is 3e-4. The learning rate begins to decrease gradually after 92,000 iterations until it finally drops to 1e-6. The GPU in experiments is V100-32GB. Some contrasting methods such as Restormer and Burstormer were retrained.

\begin{table}[!th]
\centering
  \scriptsize
  \renewcommand\arraystretch{1.1} 
	\setlength{\tabcolsep}{0.25mm}
  \caption{Quantitative comparison with the SOTA deblurring methods. The best result is in \textcolor{red}{red} and the second best one is in \textcolor{blue}{blue}. The comparison methods include DeblurGAN \cite{Deblurgan}, DeblurGAN-v2 \cite{DeblurGAN-v2}, SRN \cite{SRN}, DBGAN \cite{DBGAN}, MT-RNN \cite{MT-RNN}, DMPHN \cite{DMPHN}, MIMO-Unet+ \cite{MIMO-UNet}, MPRNet, Restormer, and Stripformer \cite{stripformer}.}
  \vspace{-0.1in}
    \begin{tabular}{l|cccccccc}
    \toprule
     \cellcolor{gray!10} &  \multicolumn{2}{c}{\cellcolor{gray!10}GoPro} & \multicolumn{2}{c}{\cellcolor{gray!10}HIDE} & \multicolumn{2}{c}{\cellcolor{gray!10}RealBlur-R } & \multicolumn{2}{c}{\cellcolor{gray!10}RealBlur-J } \\
        \cellcolor{gray!10}\multirow{-2}{*}{Methods} & \cellcolor{gray!10}PSNR$\uparrow$ & \cellcolor{gray!10}SSIM $\uparrow$ & \cellcolor{gray!10}PSNR $\uparrow$ & \cellcolor{gray!10}SSIM $\uparrow$ & \cellcolor{gray!10}PSNR $\uparrow$ & \cellcolor{gray!10}SSIM $\uparrow$ & \cellcolor{gray!10}PSNR $\uparrow$ & \cellcolor{gray!10}SSIM $\uparrow$\\
\midrule
    DeblurGAN  & 28.70  & 0.858  & 24.51  & 0.871  & 33.79  & 0.903  & 27.97  & 0.834  \\
    DeblurGAN-v2  & 29.55  & 0.934  & 26.61  & 0.875  & 35.26  & 0.944  & 28.70  & 0.866  \\
    SRN  & 30.26  & 0.934  & 28.36  & 0.915  & 35.66  & 0.947  & 28.56  & 0.867  \\
    DBGAN  & 31.10  & 0.942  & 28.94  & 0.915  & 33.78  & 0.909  & 24.93  & 0.745  \\
    MT-RNN  & 31.15  & 0.945  & 29.15  & 0.918  & 35.79  & 0.951  & 28.44  & 0.862  \\
    DMPHN  & 31.20  & 0.940  & 29.09  & 0.924  & 35.70  & 0.948  & 28.42  & 0.860  \\
    MIMO-Unet+  & 32.45  & 0.957  & 29.99  & 0.930  & 35.54  & 0.947  & 27.63  & 0.837  \\
    MPRNet  & 32.66  & 0.959  & 30.96  & 0.939  & 35.99  & 0.952  & 28.70  & 0.873  \\
    Restormer  &  32.67  & 0.960  & \textcolor{blue}{31.09}  & \textcolor{blue}{0.942}  & \textcolor{blue}{36.08}  & \textcolor{blue}{0.955}  & 28.83  & 0.875  \\
    Stripformer & \textcolor{red}{33.08}  & \textcolor{red}{0.962}  & 31.03  & 0.940  & 36.02  & 0.954  & \textcolor{blue}{28.89}  & \textcolor{blue}{0.877}  \\
    \midrule
    \textbf{EchoIR (ours)}  & \textcolor{blue}{32.82}  & \textcolor{blue}{0.962}  & \textcolor{red}{31.18}  & \textcolor{red}{0.948}  & \textcolor{red}{36.21}  & \textcolor{red}{0.958}  & \textcolor{red}{28.94}  & \textcolor{red}{0.880}  \\
    \bottomrule
    \end{tabular}%
\vspace{-0.4cm}
  \label{table:deblur_metrics}%
\end{table}

\begin{figure*}[t]
\footnotesize
\centering
\scalebox{0.92}{
\begin{tabular}{cccc}
\hspace{-0.42cm}
\begin{tabular}{c}
\includegraphics[width=0.27\textwidth]{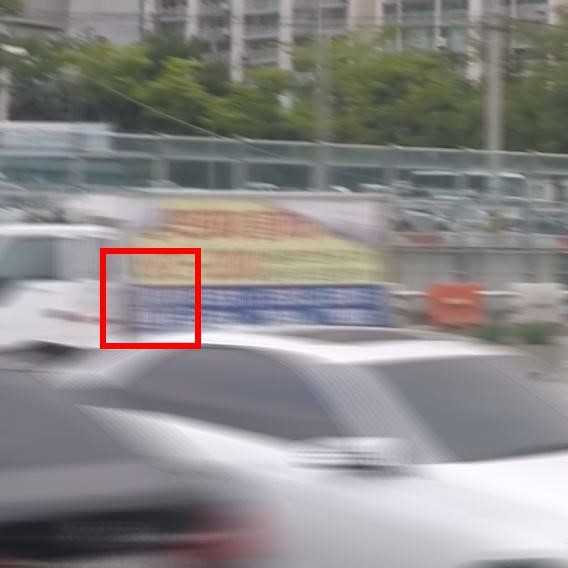}
\\
Blurry Image
\end{tabular}
\hspace{-0.1cm}
\begin{tabular}{cccccc}
\includegraphics[width=0.12\textwidth]{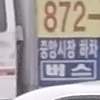} \hspace{-1mm} &
\includegraphics[width=0.12\textwidth]{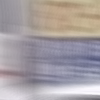} \hspace{-1mm} &
\includegraphics[width=0.12\textwidth]{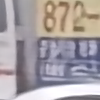} \hspace{-1mm} &
\includegraphics[width=0.12\textwidth]{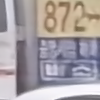} \hspace{-1mm}
\\
Reference \hspace{-1mm} &
Blurry \hspace{-1mm} &
DBGAN \hspace{-1mm} &
MIMO-Unet+ \hspace{-1mm}
\\
\includegraphics[width=0.12\textwidth]{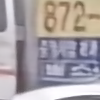} \hspace{-1mm} &
\includegraphics[width=0.12\textwidth]{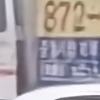} \hspace{-1mm} &
\includegraphics[width=0.12\textwidth]{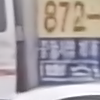} \hspace{-1mm} &
\includegraphics[width=0.12\textwidth]{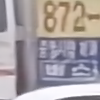} \hspace{-1mm} 
\\ 
MPRNet \hspace{-1mm} &
Restormer \hspace{-1mm} &
Stripformer \hspace{-1mm} &
Ours \hspace{-1mm} 
\\
\end{tabular}
\\

\end{tabular} }
\vspace{-0.1in}
\caption{Visual comparison of the result with SOTA deblurring methods. This comparison employs the output results for direct comparison, thereby providing a clear reflection of the quality of the model's output outcomes.}
\label{fig:deblur_visual}
\end{figure*}

\subsection{Deraining Results}
The experiment validates the training outcomes of our model using four benchmark datasets: Test100 \cite{Test100}, Rain100H \cite{Rain100}, Rain100L, Test2800 \cite{Test2800}, and Test1200 \cite{Test1200}. As depicted in Table \ref{table:derain_metrics}, our model exhibits remarkable performance across all datasets. Among all datasets, our model demonstrates the highest PSNR performance on the Test100 dataset, surpassing the second-best model by 0.45 dB. Additionally, it achieves the best SSIM performance on the Test1200 dataset, exceeding the second place by 0.015.

The experiment also entailed a visual comparison of the deraining task, as depicted in Figure \ref{fig:derainning_visual}. DGUNet and MPRNet struggle to effectively remove rain and fog. While MFDNet and PReNet manage to eliminate most rain, residual noise from the precipitation remains in some images. Restormer, while successful in rain removal, exhibits noticeable texture and color loss, particularly on the door surface.  The results of EchoIR closely resemble ground truth, with rain cleanly removed, showcasing the effectiveness of EchoIR in rain removal tasks.

\subsection{Deblurring Results}
The experiment validates the training outcomes of our model using four benchmark datasets: GoPro \cite{GoPro}, HIDE \cite{HIDE}, RealBlur-R \cite{RealBlur}, and RealBlur-J \cite{RealBlur}. As depicted in Table \ref{table:deblur_metrics}, our model exhibits remarkable performance across all datasets. With the exception of the performance on the GoPro dataset, where it lags behind Stripformer, EchoIR exhibits the highest PSNR and SSIM performance. Particularly noteworthy is our model's superior performance on the RealBlur-R dataset, with PSNR and SSIM values exceeding the second-highest results by 0.13dB and 0.003, respectively. This underscores the efficacy of our model's feature extraction capabilities in the realm of image restoration.

The experiment further assessed the visual performance of the model results to facilitate a more intuitive comparison between models. Figure \ref{fig:deblur_visual} illustrates the image details of the model output. In the original blurry image, the text on the truck compartment is notably indistinct and barely discernible, while the boundary between the white car's roof and the truck appears blurred as well. In the restored results, various methods exhibit different degrees of restoration. For instance, MIMO-Unet+ effectively enhances the clarity of the text but struggles with the blurriness between the car roof and the text. On the other hand, models like Restormer and Stripformer manage to restore the relationship between the car roof and the text, yet the text remains somewhat blurred. In contrast, our model excels in both aspects simultaneously. This indicates that our model demonstrates a more comprehensive feature extraction capability, allowing it to effectively utilize features for image restoration.

\begin{table*}[!th]
\centering
  \scriptsize
  \renewcommand\arraystretch{1.5} 
	\setlength{\tabcolsep}{0.7mm}
  \caption{Metrics comparison with the SOTA denoising methods. The best result is in \textcolor{red}{red} and the second best one is in \textcolor{blue}{blue}. The compared methods include RIDNet \cite{RIDNet}, VDN \cite{VDN}, SADNet \cite{SADNet}, DANet+ \cite{DANet}, CycleISP \cite{cycleisp}, MIRNet \cite{MIRNet}, DreamNet \cite{DreamNet}, MPRNet, DAGL \cite{DAGL}, SwinIR, Uformer \cite{Uformer}, Restormer, AirNet \cite{AirNet}, and Burstormer.}
  \vspace{-0.1in}
    \begin{tabular}{l|c|ccccccccccccccc}
    \toprule
        \cellcolor{gray!10}Datasets & \cellcolor{gray!10}Metrics & \cellcolor{gray!10}RIDNet & \cellcolor{gray!10}VDN & \cellcolor{gray!10}SADNet & \cellcolor{gray!10}DANet+ & \cellcolor{gray!10}CycleISP & \cellcolor{gray!10}MIRNet & \cellcolor{gray!10}DeamNet & \cellcolor{gray!10}MPRNet & \cellcolor{gray!10}DAGL & \cellcolor{gray!10}SwinIR & \cellcolor{gray!10}Uformer & \cellcolor{gray!10}Restormer & \cellcolor{gray!10}AirNet & \cellcolor{gray!10}Burstormer & \cellcolor{gray!10}\textbf{EchoIR}\\
\midrule
     & PSNR$\uparrow$  & 38.71  & 39.28  & 39.46  & 39.47  & 39.52  & 39.72  & 39.47  & 39.71  & 38.94  & 39.81 & 39.77  & 39.91  & \textcolor{blue}{39.95}  & 39.93  & \textcolor{red}{40.05}  \\
    \multirow{-2}{*}{SIDD} & SSIM $\uparrow$  & 0.951  & 0.956  & 0.957  & 0.957  & 0.957  & \textcolor{blue}{0.959}  & 0.957  & 0.958  & 0.953  & 0.958  & \textcolor{blue}{0.959}  & 0.958  & 0.958  & 0.958  & \textcolor{red}{0.962}  \\
    \midrule
     & PSNR$\uparrow$  & 39.26  & 39.38  & 39.59  & 39.58  & 39.56  & 39.88  & 39.63  & 39.80  & 39.77  & 39.91  & 39.96  & 39.97  & \textcolor{blue}{39.99}  & 39.96  & \textcolor{red}{40.06}  \\
    \multirow{-2}{*}{DND} & SSIM $\uparrow$  & 0.953  & 0.952  & 0.952  & 0.955  & \textcolor{blue}{0.956}  & \textcolor{blue}{0.956}  & 0.953  & 0.954  & \textcolor{blue}{0.956}  & 0.955  & \textcolor{blue}{0.956}  & \textcolor{blue}{0.956}  & 0.955  & 0.955  & \textcolor{red}{0.960}  \\
    \bottomrule
    \end{tabular}%
  \label{table:denoise_metrics}%
\end{table*}

\begin{figure*}[!th]
\footnotesize
\centering
\renewcommand\arraystretch{1.5} 
\setlength{\tabcolsep}{0.5pt}
\begin{tabular}{cccccccc}
\includegraphics[width=0.1\textwidth]{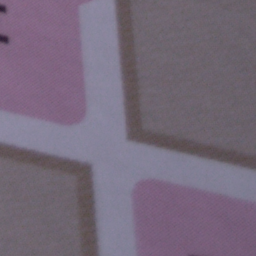} &
\includegraphics[width=0.1\textwidth]{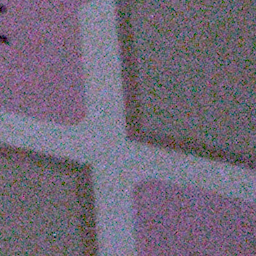} &
\includegraphics[width=0.1\textwidth]{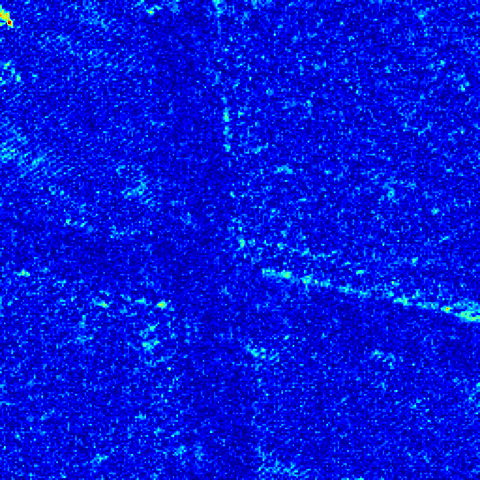} &
\includegraphics[width=0.1\textwidth]{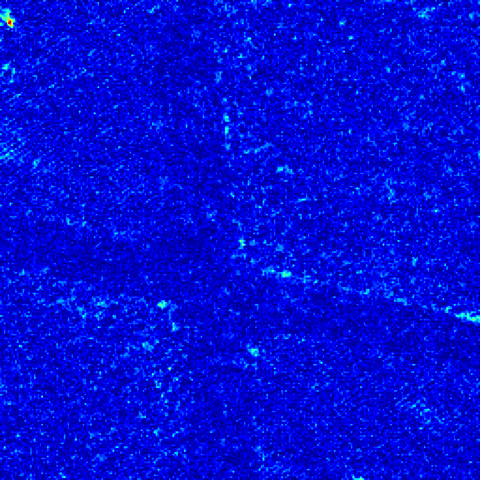} &
\includegraphics[width=0.1\textwidth]{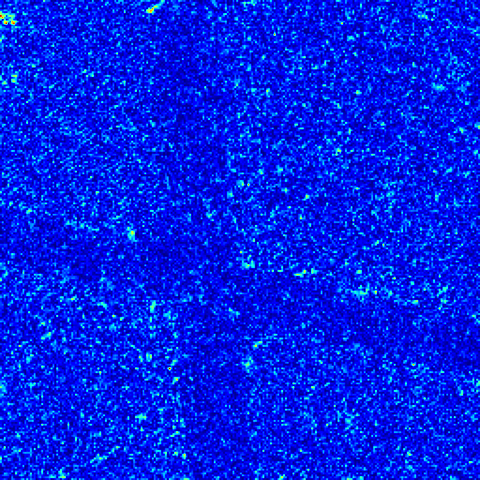} &
\includegraphics[width=0.1\textwidth]{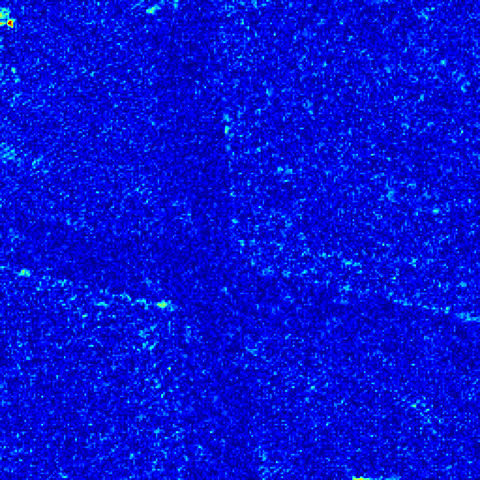} &
\includegraphics[width=0.1\textwidth]{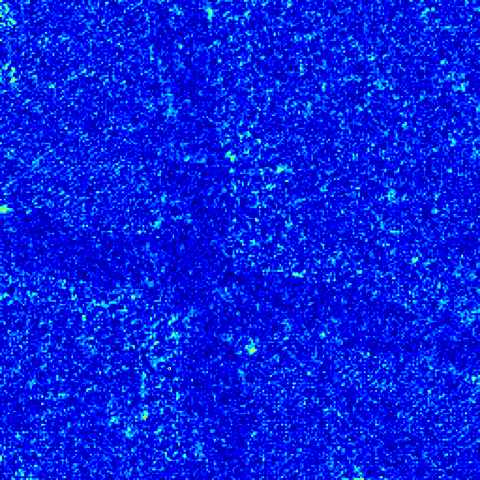} &
\includegraphics[width=0.1\textwidth]{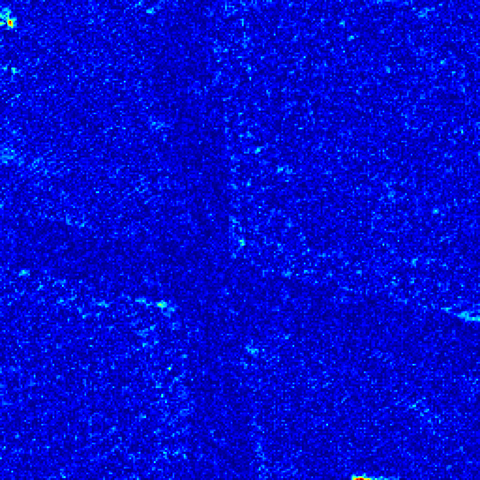}
\\
\includegraphics[width=0.1\textwidth]{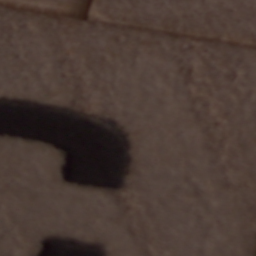} &
\includegraphics[width=0.1\textwidth]{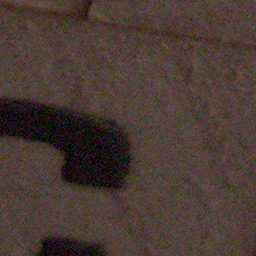} &
\includegraphics[width=0.1\textwidth]{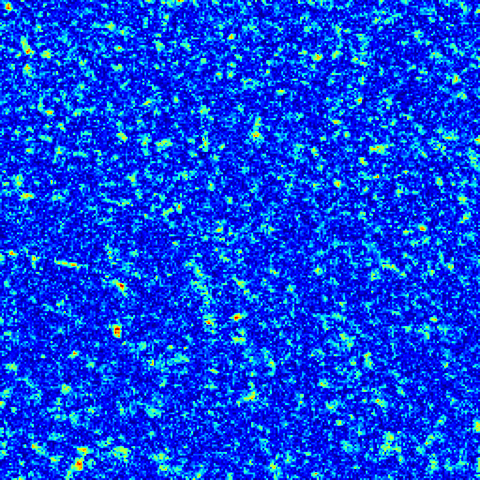} &
\includegraphics[width=0.1\textwidth]{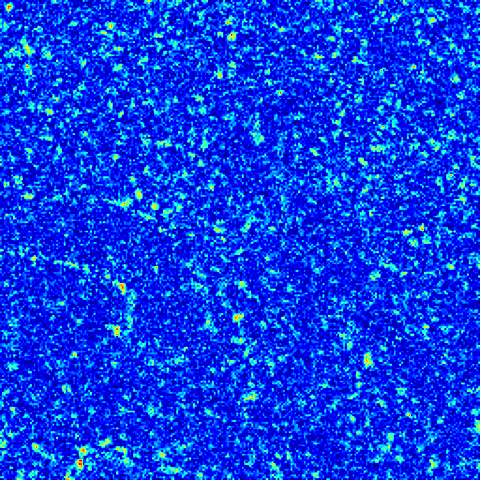} &
\includegraphics[width=0.1\textwidth]{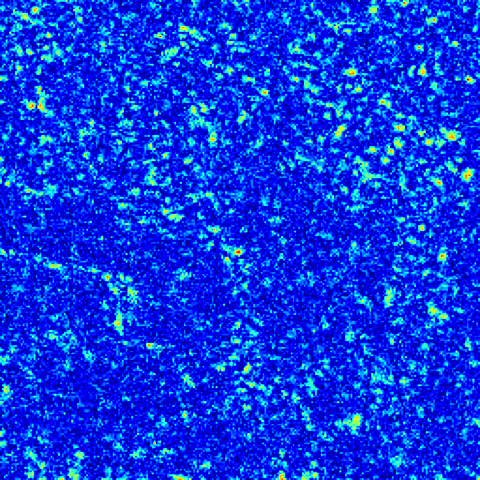} &
\includegraphics[width=0.1\textwidth]{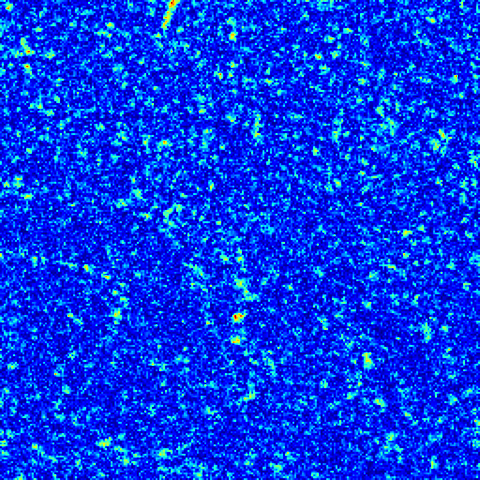} &
\includegraphics[width=0.1\textwidth]{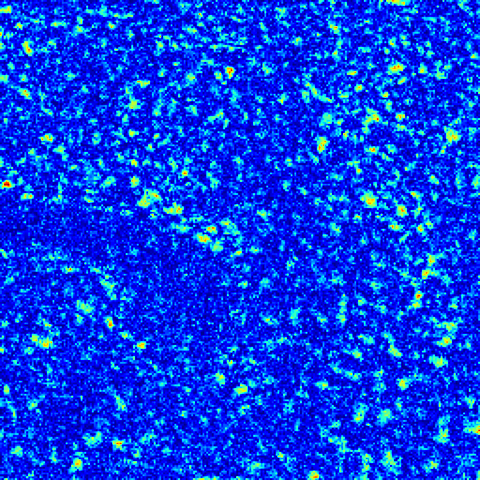} &
\includegraphics[width=0.1\textwidth]{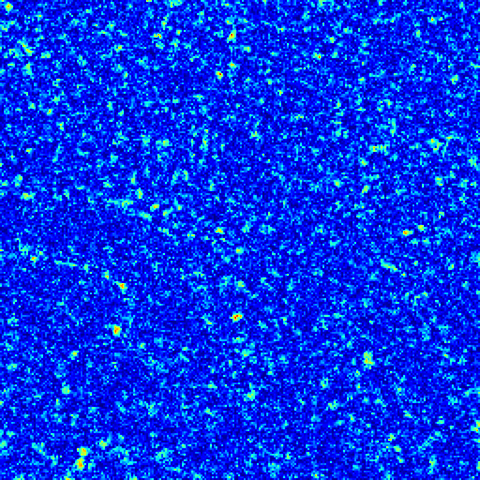}
\\
Gt &
Noise &
MIRNet  &
MPRNet  &
SwinIR  &
Restormer &
AirNet &
Ours 
\end{tabular}
\vspace{-0.1in}
\caption{Visual comparison of the result with SOTA denosing methods. This comparison employs the difference map for clear comparison, thereby providing a clear reflection of the quality of the model's output outcomes.}
\label{fig:denoise_visual}
\end{figure*}

\begin{table}[!t]
        \centering
        \setlength{\tabcolsep}{3.2mm}
  \caption{Ablation study of Echo-Upsampler. PS stands for simple Pixel Shuffle method. TC stands for Transpose Convolution. EU stands for Echo-Upsampler referencing feature maps. The results are the average values of five test datasets.}
  \vspace{-0.1in}
    \begin{tabular}{lccccc}
    \toprule
    \rowcolor{gray!10} Type & Params(M) & FLOPs(G) & PSNR  & SSIM \\
    \midrule
     PS     & 33,43 & 183.33 & 31.18 & 0.895 \\
     TC    & 30.25 & 178.69 & 31.24 & 0.901 \\
     EU    & 30.28 & 176.35 & 31.43 & 0.908 \\
    \bottomrule
    \end{tabular}%
  \label{EchoUpsampler Ablation}%
\vspace{-5pt}
\end{table}

\begin{table}[!t]
        \centering
        \setlength{\tabcolsep}{1.5mm}
  \caption{Ablation study of Mix-Attention Block. A check mark indicates that this structure is adopted.}
  \vspace{-0.1in}
    \begin{tabular}{lcccccc}
    \toprule
    \rowcolor{gray!10} CA & GDFN & Params(M) & FLOPs(G) & PSNR  & SSIM \\
    \midrule
        &  & 13.12 & 85.43 & 36.35 & 0.930 \\
        & \checkmark & 22.98 & 176.31 & 39.53 & 0.953 \\
        \checkmark &  & 20.42 & 85.46 & 36.56 & 0.932 \\
        \checkmark & \checkmark & 30.26 & 176.35 & 39.61 & 0.955 \\
    \bottomrule
    \end{tabular}%
  \label{Attention Ablation}%
\vspace{-10pt}
\end{table}

\vspace{-5pt}
\subsection{Denoising Results}
In the denoising experiment, emphasis was placed on real image denoising tasks, with verification conducted on datasets such as SIDD \cite{SIDD} and DND \cite{DND}. As demonstrated in Table \ref{table:denoise_metrics}, our method exhibits exceptional metric performance across both datasets. This indicates that the images generated by our model closely resemble the ground truth results, highlighting the model's accuracy in noise removal and information restoration tasks in denoising. 

For the visual comparison of the denoising task, we utilize the difference map between the denoising results and the ground truth to enhance the contrast effect. In this representation, more blue areas signify smaller differences, while more green or red areas indicate greater disparities. As illustrated in the results depicted in Figure \ref{fig:denoise_visual}, other methods exhibit a noticeable presence of green areas in the comparison between the two groups. Conversely, our method outperforms others in both sets of difference maps, with the blue portion occupying the largest proportion and appearing the deepest. This suggests that the denoising results generated by our model closely resemble the ground truth. Additionally, this underscores the efficacy of our model's feature extraction approach in denoising tasks.


\subsection{Ablation Study}
The experiment setting of the ablation study is the same as the complete experiment except the number of iterations was reduced to 1e+5. Experiments compare PSNR, SIMM, and FLOPs. Ablation experiments examine the necessity of the Echo Upsampler, AS-BLO, and Mix-Attention blocks.

\textbf{Echo-Upsampler.} The task for the ablation experiment of the Echo-Upsampler is single-image deraining. We first replace it with a simple upsampler to compare the effect. A simple upsampler consists of a convolutional layer and a pixel shuffle. For the test of the Echo-Upsampler, we changed its reference object to the input image $I_{L}$ instead of feature maps from the encoder. The experimental results are shown in Table \ref{EchoUpsampler Ablation}. From the experimental results, the best results can be achieved by using an Echo-Upsampler and using encoder feature maps of different levels as reference input.

\textbf{Mix-Attention block.} The task of the ablation experiment of the Mix-Attention block is denoising. In the experiment of the Mix-Attention block, we first ablated the channel attention part and only retained the multi-head self-attention in the attention calculation. We then ablated GDFN and replaced FFN with a multi-layer perceptron composed of two linear projections and GeLU activation functions. The experimental results are shown in Table \ref{Attention Ablation}.

\textbf{Training Strategy.} An optimized alternative to AS-BLO is AdamW, with parameters $b_1=0.9$, $b_2=0.999$, and weight decay is $1e-4$. The ablation study of training strategies was performed on three tasks. Table.~\ref{BLO Ablation} showcases the enhancements achieved using our approximated sequential bi-level optimization approach relative to the direct joint training with our proposed EchoIR network.
\begin{table}[!th]
        \centering
        \setlength{\tabcolsep}{2.4mm}
  \caption{Ablation study of bi-level optimization strategy. SL is the single-level optimization strategy. The table displays the average PSNR of the Deraining, Deblurring, and Denoising tasks on different datasets}
  \vspace{-0.1in}
    \begin{tabular}{lcccc}
    \toprule
    \rowcolor{gray!10} Type & FLOPs(G) & Derain  & Deblur  & Denoise \\
    \midrule
     SL     & 171.41 & 31.21 & 30.25 & 39.36 \\
     AS-BLO    & 176.35 & 31.43 & 30.61 & 39.61 \\
    \bottomrule
    \end{tabular}%
  \label{BLO Ablation}%
\vspace{-10pt}
\end{table} Our strategy posits that EchoIR should yield lossless image upsampling that is conducive to both human visual assessment and computational perception. By approximating the complex bi-level optimization into a sequence of single-level optimization tasks solvable by gradient descent, we reach a notably higher PSNR and SSIM.

\section{Conclusion}
In this paper, we propose the EchoIR model for various image restoration tasks. This model features an Echo-Upsampler, which utilizes encoder-provided feature maps for efficient upsampling despite poor input smoothness. To enhance feature extraction, we incorporate a Mix-Attention module. We employ the AS-BLO training strategy to fully train the Echo-Upsampler and achieve better reconstruction, optimizing the model during training. This strategy effectively models hierarchical tasks in image restoration and upsampling. Experiments across multiple datasets and tasks show that our model achieves state-of-the-art performance in various image restoration tasks.
\textbf{Broader Impacts.} EchoIR innovatively divides the entire process into the upper-layer image restoration task and the lower-layer upsampling task, and uses Approximated Sequential Bi-level Optimization for optimization. EchoIR provides a new paradigm for image restoration tasks.

\bibliography{aaai25}

\begin{thebibliography}{64}
\providecommand{\natexlab}[1]{#1}

\bibitem[{Abdelhamed, Lin, and Brown(2018)}]{SIDD}
Abdelhamed, A.; Lin, S.; and Brown, M.~S. 2018.
\newblock A high-quality denoising dataset for smartphone cameras.
\newblock In \emph{Proceedings of the IEEE conference on computer vision and pattern recognition}, 1692--1700.

\bibitem[{Anwar and Barnes(2019)}]{RIDNet}
Anwar, S.; and Barnes, N. 2019.
\newblock Real image denoising with feature attention.
\newblock In \emph{Proceedings of the IEEE/CVF international conference on computer vision}, 3155--3164.

\bibitem[{Cavigelli, Hager, and Benini(2017)}]{CASCNN}
Cavigelli, L.; Hager, P.; and Benini, L. 2017.
\newblock CAS-CNN: A deep convolutional neural network for image compression artifact suppression.
\newblock In \emph{2017 International Joint Conference on Neural Networks (IJCNN)}, 752--759.

\bibitem[{Chang et~al.(2020)Chang, Li, Feng, and Xu}]{SADNet}
Chang, M.; Li, Q.; Feng, H.; and Xu, Z. 2020.
\newblock Spatial-adaptive network for single image denoising.
\newblock In \emph{Computer Vision--ECCV 2020: 16th European Conference, Glasgow, UK, August 23--28, 2020, Proceedings, Part XXX 16}, 171--187. Springer.

\bibitem[{Chen et~al.(2021)Chen, Wang, Guo, Xu, Deng, Liu, Ma, Xu, Xu, and Gao}]{IPT}
Chen, H.; Wang, Y.; Guo, T.; Xu, C.; Deng, Y.; Liu, Z.; Ma, S.; Xu, C.; Xu, C.; and Gao, W. 2021.
\newblock Pre-Trained Image Processing Transformer.
\newblock arXiv:2012.00364.

\bibitem[{Chen et~al.(2023{\natexlab{a}})Chen, Wang, Zhou, Qiao, and Dong}]{chen2023activating}
Chen, X.; Wang, X.; Zhou, J.; Qiao, Y.; and Dong, C. 2023{\natexlab{a}}.
\newblock Activating More Pixels in Image Super-Resolution Transformer.
\newblock In \emph{Proceedings of the IEEE/CVF Conference on Computer Vision and Pattern Recognition (CVPR)}, 22367--22377.

\bibitem[{Chen et~al.(2024)Chen, Zhang, Gu, Kong, and Yang}]{chen2024recursive}
Chen, Z.; Zhang, Y.; Gu, J.; Kong, L.; and Yang, X. 2024.
\newblock Recursive Generalization Transformer for Image Super-Resolution.
\newblock In \emph{ICLR}.

\bibitem[{Chen et~al.(2023{\natexlab{b}})Chen, Zhang, Gu, Kong, Yang, and Yu}]{DAT}
Chen, Z.; Zhang, Y.; Gu, J.; Kong, L.; Yang, X.; and Yu, F. 2023{\natexlab{b}}.
\newblock Dual aggregation transformer for image super-resolution.
\newblock In \emph{Proceedings of the IEEE/CVF international conference on computer vision}, 12312--12321.

\bibitem[{Chen et~al.(2023{\natexlab{c}})Chen, Zhang, Gu, Kong, Yang, and Yu}]{chen2023dual}
Chen, Z.; Zhang, Y.; Gu, J.; Kong, L.; Yang, X.; and Yu, F. 2023{\natexlab{c}}.
\newblock Dual Aggregation Transformer for Image Super-Resolution.
\newblock In \emph{ICCV}.

\bibitem[{Cho et~al.(2021)Cho, Ji, Hong, Jung, and Ko}]{MIMO-UNet}
Cho, S.-J.; Ji, S.-W.; Hong, J.-P.; Jung, S.-W.; and Ko, S.-J. 2021.
\newblock Rethinking coarse-to-fine approach in single image deblurring.
\newblock In \emph{Proceedings of the IEEE/CVF international conference on computer vision}, 4641--4650.

\bibitem[{Dong et~al.(2014)Dong, Loy, He, and Tang}]{SRCNN}
Dong, C.; Loy, C.~C.; He, K.; and Tang, X. 2014.
\newblock Image Super-Resolution Using Deep Convolutional Networks.
\newblock \emph{IEEE Transactions on Pattern Analysis and Machine Intelligence}, 38.

\bibitem[{Dosovitskiy et~al.(2021)Dosovitskiy, Beyer, Kolesnikov, Weissenborn, Zhai, Unterthiner, Dehghani, Minderer, Heigold, Gelly, Uszkoreit, and Houlsby}]{ViT}
Dosovitskiy, A.; Beyer, L.; Kolesnikov, A.; Weissenborn, D.; Zhai, X.; Unterthiner, T.; Dehghani, M.; Minderer, M.; Heigold, G.; Gelly, S.; Uszkoreit, J.; and Houlsby, N. 2021.
\newblock An Image is Worth 16x16 Words: Transformers for Image Recognition at Scale.
\newblock \emph{ICLR}.

\bibitem[{Dudhane et~al.(2023)Dudhane, Zamir, Khan, Khan, and Yang}]{Burstormer}
Dudhane, A.; Zamir, S.~W.; Khan, S.; Khan, F.~S.; and Yang, M.-H. 2023.
\newblock Burstormer: Burst image restoration and enhancement transformer.
\newblock In \emph{2023 IEEE/CVF Conference on Computer Vision and Pattern Recognition (CVPR)}, 5703--5712. IEEE.

\bibitem[{Fu et~al.(2017{\natexlab{a}})Fu, Huang, Ding, Liao, and Paisley}]{DerainNet}
Fu, X.; Huang, J.; Ding, X.; Liao, Y.; and Paisley, J. 2017{\natexlab{a}}.
\newblock Clearing the skies: A deep network architecture for single-image rain removal.
\newblock \emph{IEEE Transactions on Image Processing}, 26(6): 2944--2956.

\bibitem[{Fu et~al.(2017{\natexlab{b}})Fu, Huang, Zeng, Huang, Ding, and Paisley}]{Test2800}
Fu, X.; Huang, J.; Zeng, D.; Huang, Y.; Ding, X.; and Paisley, J. 2017{\natexlab{b}}.
\newblock Removing rain from single images via a deep detail network.
\newblock In \emph{Proceedings of the IEEE conference on computer vision and pattern recognition}, 3855--3863.

\bibitem[{Gulrajani et~al.(2017)Gulrajani, Ahmed, Arjovsky, Dumoulin, and Courville}]{NIPS2017_892c3b1c}
Gulrajani, I.; Ahmed, F.; Arjovsky, M.; Dumoulin, V.; and Courville, A.~C. 2017.
\newblock Improved Training of Wasserstein GANs.
\newblock In Guyon, I.; Luxburg, U.~V.; Bengio, S.; Wallach, H.; Fergus, R.; Vishwanathan, S.; and Garnett, R., eds., \emph{Advances in Neural Information Processing Systems}, volume~30. Curran Associates, Inc.

\bibitem[{He et~al.(2016)He, Zhang, Ren, and Sun}]{ResNet}
He, K.; Zhang, X.; Ren, S.; and Sun, J. 2016.
\newblock Deep residual learning for image recognition.
\newblock In \emph{Proceedings of the IEEE conference on computer vision and pattern recognition}, 770--778.

\bibitem[{Jiang et~al.(2020)Jiang, Wang, Yi, Chen, Huang, Luo, Ma, and Jiang}]{MSPFN}
Jiang, K.; Wang, Z.; Yi, P.; Chen, C.; Huang, B.; Luo, Y.; Ma, J.; and Jiang, J. 2020.
\newblock Multi-scale progressive fusion network for single image deraining.
\newblock In \emph{Proceedings of the IEEE/CVF conference on computer vision and pattern recognition}, 8346--8355.

\bibitem[{Jiang et~al.(2024)Jiang, Li, Liu, Fan, and Liu}]{jiang2024towards}
Jiang, Z.; Li, X.; Liu, J.; Fan, X.; and Liu, R. 2024.
\newblock Towards Robust Image Stitching: An Adaptive Resistance Learning against Compatible Attacks.
\newblock In \emph{Proceedings of the AAAI Conference on Artificial Intelligence}, volume~38, 2589--2597.

\bibitem[{Kim, Lee, and Lee(2016)}]{Kim_2016_CVPR}
Kim, J.; Lee, J.~K.; and Lee, K.~M. 2016.
\newblock Accurate Image Super-Resolution Using Very Deep Convolutional Networks.
\newblock In \emph{Proceedings of the IEEE Conference on Computer Vision and Pattern Recognition (CVPR)}.

\bibitem[{Kopf et~al.(2007)Kopf, Cohen, Lischinski, and Uyttendaele}]{JBU}
Kopf, J.; Cohen, M.~F.; Lischinski, D.; and Uyttendaele, M. 2007.
\newblock Joint bilateral upsampling.
\newblock \emph{ACM Transactions on Graphics (ToG)}, 26(3): 96--es.

\bibitem[{Kupyn et~al.(2018)Kupyn, Budzan, Mykhailych, Mishkin, and Matas}]{Deblurgan}
Kupyn, O.; Budzan, V.; Mykhailych, M.; Mishkin, D.; and Matas, J. 2018.
\newblock Deblurgan: Blind motion deblurring using conditional adversarial networks.
\newblock In \emph{Proceedings of the IEEE conference on computer vision and pattern recognition}, 8183--8192.

\bibitem[{Kupyn et~al.(2019)Kupyn, Martyniuk, Wu, and Wang}]{DeblurGAN-v2}
Kupyn, O.; Martyniuk, T.; Wu, J.; and Wang, Z. 2019.
\newblock Deblurgan-v2: Deblurring (orders-of-magnitude) faster and better.
\newblock In \emph{Proceedings of the IEEE/CVF international conference on computer vision}, 8878--8887.

\bibitem[{LeCun, Bengio, and Hinton(2015)}]{CNN}
LeCun, Y.; Bengio, Y.; and Hinton, G. 2015.
\newblock Deep learning.
\newblock \emph{nature}, 521(7553): 436--444.

\bibitem[{Li et~al.(2022{\natexlab{a}})Li, Liu, Hu, Wu, Lv, and Peng}]{AirNet}
Li, B.; Liu, X.; Hu, P.; Wu, Z.; Lv, J.; and Peng, X. 2022{\natexlab{a}}.
\newblock All-in-one image restoration for unknown corruption.
\newblock In \emph{Proceedings of the IEEE/CVF Conference on Computer Vision and Pattern Recognition}, 17452--17462.

\bibitem[{Li et~al.(2022{\natexlab{b}})Li, Lu, Qian, Lu, Zhang, and Jia}]{li2022efficient}
Li, W.; Lu, X.; Qian, S.; Lu, J.; Zhang, X.; and Jia, J. 2022{\natexlab{b}}.
\newblock On Efficient Transformer-Based Image Pre-training for Low-Level Vision.
\newblock arXiv:2112.10175.

\bibitem[{Li et~al.(2025)Li, Liu, Chen, Zou, Ma, Fan, and Liu}]{li2025contourlet}
Li, X.; Liu, J.; Chen, Z.; Zou, Y.; Ma, L.; Fan, X.; and Liu, R. 2025.
\newblock Contourlet residual for prompt learning enhanced infrared image super-resolution.
\newblock In \emph{European Conference on Computer Vision}, 270--288. Springer.

\bibitem[{Li et~al.(2023)Li, Zou, Liu, Jiang, Ma, Fan, and Liu}]{li2023text}
Li, X.; Zou, Y.; Liu, J.; Jiang, Z.; Ma, L.; Fan, X.; and Liu, R. 2023.
\newblock From Text to Pixels: A Context-Aware Semantic Synergy Solution for Infrared and Visible Image Fusion.
\newblock \emph{arXiv preprint arXiv:2401.00421}.

\bibitem[{Liang et~al.(2021)Liang, Cao, Sun, Zhang, Van~Gool, and Timofte}]{SwinIR}
Liang, J.; Cao, J.; Sun, G.; Zhang, K.; Van~Gool, L.; and Timofte, R. 2021.
\newblock SwinIR: Image Restoration Using Swin Transformer.
\newblock \emph{arXiv preprint arXiv:2108.10257}.

\bibitem[{Mou, Wang, and Zhang(2022)}]{DGUNet}
Mou, C.; Wang, Q.; and Zhang, J. 2022.
\newblock Deep generalized unfolding networks for image restoration.
\newblock In \emph{Proceedings of the IEEE/CVF Conference on Computer Vision and Pattern Recognition}, 17399--17410.

\bibitem[{Mou, Zhang, and Wu(2021)}]{DAGL}
Mou, C.; Zhang, J.; and Wu, Z. 2021.
\newblock Dynamic attentive graph learning for image restoration.
\newblock In \emph{Proceedings of the IEEE/CVF international conference on computer vision}, 4328--4337.

\bibitem[{Nah, Hyun~Kim, and Mu~Lee(2017)}]{GoPro}
Nah, S.; Hyun~Kim, T.; and Mu~Lee, K. 2017.
\newblock Deep multi-scale convolutional neural network for dynamic scene deblurring.
\newblock In \emph{Proceedings of the IEEE conference on computer vision and pattern recognition}, 3883--3891.

\bibitem[{Park et~al.(2020)Park, Kang, Kim, and Chun}]{MT-RNN}
Park, D.; Kang, D.~U.; Kim, J.; and Chun, S.~Y. 2020.
\newblock Multi-temporal recurrent neural networks for progressive non-uniform single image deblurring with incremental temporal training.
\newblock In \emph{European Conference on Computer Vision}, 327--343. Springer.

\bibitem[{Plotz and Roth(2017)}]{DND}
Plotz, T.; and Roth, S. 2017.
\newblock Benchmarking denoising algorithms with real photographs.
\newblock In \emph{Proceedings of the IEEE conference on computer vision and pattern recognition}, 1586--1595.

\bibitem[{Purohit et~al.(2021)Purohit, Suin, Rajagopalan, and Boddeti}]{SPAIR}
Purohit, K.; Suin, M.; Rajagopalan, A.; and Boddeti, V.~N. 2021.
\newblock Spatially-adaptive image restoration using distortion-guided networks.
\newblock In \emph{Proceedings of the IEEE/CVF international conference on computer vision}, 2309--2319.

\bibitem[{Ren et~al.(2021)Ren, He, Wang, and Zhao}]{DreamNet}
Ren, C.; He, X.; Wang, C.; and Zhao, Z. 2021.
\newblock Adaptive consistency prior based deep network for image denoising.
\newblock In \emph{Proceedings of the IEEE/CVF conference on computer vision and pattern recognition}, 8596--8606.

\bibitem[{Ren et~al.(2019)Ren, Zuo, Hu, Zhu, and Meng}]{PReNet}
Ren, D.; Zuo, W.; Hu, Q.; Zhu, P.; and Meng, D. 2019.
\newblock Progressive image deraining networks: A better and simpler baseline.
\newblock In \emph{Proceedings of the IEEE/CVF conference on computer vision and pattern recognition}, 3937--3946.

\bibitem[{Rim et~al.(2020)Rim, Lee, Won, and Cho}]{RealBlur}
Rim, J.; Lee, H.; Won, J.; and Cho, S. 2020.
\newblock Real-world blur dataset for learning and benchmarking deblurring algorithms.
\newblock In \emph{Computer Vision--ECCV 2020: 16th European Conference, Glasgow, UK, August 23--28, 2020, Proceedings, Part XXV 16}, 184--201. Springer.

\bibitem[{Ronneberger, Fischer, and Brox(2015)}]{U-Net}
Ronneberger, O.; Fischer, P.; and Brox, T. 2015.
\newblock U-net: Convolutional networks for biomedical image segmentation.
\newblock In \emph{Medical image computing and computer-assisted intervention--MICCAI 2015: 18th international conference, Munich, Germany, October 5-9, 2015, proceedings, part III 18}, 234--241. Springer.

\bibitem[{Shen et~al.(2019)Shen, Wang, Lu, Shen, Ling, Xu, and Shao}]{HIDE}
Shen, Z.; Wang, W.; Lu, X.; Shen, J.; Ling, H.; Xu, T.; and Shao, L. 2019.
\newblock Human-aware motion deblurring.
\newblock In \emph{Proceedings of the IEEE/CVF International Conference on Computer Vision}, 5572--5581.

\bibitem[{Tao et~al.(2018)Tao, Gao, Shen, Wang, and Jia}]{SRN}
Tao, X.; Gao, H.; Shen, X.; Wang, J.; and Jia, J. 2018.
\newblock Scale-recurrent network for deep image deblurring.
\newblock In \emph{Proceedings of the IEEE conference on computer vision and pattern recognition}, 8174--8182.

\bibitem[{Tsai et~al.(2022)Tsai, Peng, Lin, Tsai, and Lin}]{stripformer}
Tsai, F.-J.; Peng, Y.-T.; Lin, Y.-Y.; Tsai, C.-C.; and Lin, C.-W. 2022.
\newblock Stripformer: Strip transformer for fast image deblurring.
\newblock In \emph{European Conference on Computer Vision}, 146--162. Springer.

\bibitem[{Vaswani et~al.(2017)Vaswani, Shazeer, Parmar, Uszkoreit, Jones, Gomez, Kaiser, and Polosukhin}]{Transformer}
Vaswani, A.; Shazeer, N.; Parmar, N.; Uszkoreit, J.; Jones, L.; Gomez, A.~N.; Kaiser, L.~u.; and Polosukhin, I. 2017.
\newblock Attention is All you Need.
\newblock In Guyon, I.; Luxburg, U.~V.; Bengio, S.; Wallach, H.; Fergus, R.; Vishwanathan, S.; and Garnett, R., eds., \emph{Advances in Neural Information Processing Systems}, volume~30. Curran Associates, Inc.

\bibitem[{Wang et~al.(2023)Wang, Jiang, Wang, Ren, Zhang, and Lin}]{MFDNet}
Wang, Q.; Jiang, K.; Wang, Z.; Ren, W.; Zhang, J.; and Lin, C.-W. 2023.
\newblock Multi-scale fusion and decomposition network for single image deraining.
\newblock \emph{IEEE Transactions on Image Processing}, 33: 191--204.

\bibitem[{Wang et~al.(2018)Wang, Yu, Wu, Gu, Liu, Dong, Qiao, and Change~Loy}]{Wang_2018_ECCV_Workshops}
Wang, X.; Yu, K.; Wu, S.; Gu, J.; Liu, Y.; Dong, C.; Qiao, Y.; and Change~Loy, C. 2018.
\newblock ESRGAN: Enhanced Super-Resolution Generative Adversarial Networks.
\newblock In \emph{Proceedings of the European Conference on Computer Vision (ECCV) Workshops}.

\bibitem[{Wang et~al.(2022)Wang, Cun, Bao, Zhou, Liu, and Li}]{Uformer}
Wang, Z.; Cun, X.; Bao, J.; Zhou, W.; Liu, J.; and Li, H. 2022.
\newblock Uformer: A General U-Shaped Transformer for Image Restoration.
\newblock In \emph{Proceedings of the IEEE/CVF Conference on Computer Vision and Pattern Recognition (CVPR)}, 17683--17693.

\bibitem[{Wei et~al.(2019)Wei, Meng, Zhao, Xu, and Wu}]{semi}
Wei, W.; Meng, D.; Zhao, Q.; Xu, Z.; and Wu, Y. 2019.
\newblock Semi-supervised transfer learning for image rain removal.
\newblock In \emph{Proceedings of the IEEE/CVF conference on computer vision and pattern recognition}, 3877--3886.

\bibitem[{Yang et~al.(2017)Yang, Tan, Feng, Liu, Guo, and Yan}]{Rain100}
Yang, W.; Tan, R.~T.; Feng, J.; Liu, J.; Guo, Z.; and Yan, S. 2017.
\newblock Deep joint rain detection and removal from a single image.
\newblock In \emph{Proceedings of the IEEE conference on computer vision and pattern recognition}, 1357--1366.

\bibitem[{Yu et~al.(2016)Yu, Dong, Loy, and Tang}]{ARCNN}
Yu, K.; Dong, C.; Loy, C.~C.; and Tang, X. 2016.
\newblock Deep Convolution Networks for Compression Artifacts Reduction.
\newblock arXiv:1608.02778.

\bibitem[{Yue et~al.(2019)Yue, Yong, Zhao, Meng, and Zhang}]{VDN}
Yue, Z.; Yong, H.; Zhao, Q.; Meng, D.; and Zhang, L. 2019.
\newblock Variational denoising network: Toward blind noise modeling and removal.
\newblock \emph{Advances in neural information processing systems}, 32.

\bibitem[{Yue et~al.(2020)Yue, Zhao, Zhang, and Meng}]{DANet}
Yue, Z.; Zhao, Q.; Zhang, L.; and Meng, D. 2020.
\newblock Dual adversarial network: Toward real-world noise removal and noise generation.
\newblock In \emph{Computer Vision--ECCV 2020: 16th European Conference, Glasgow, UK, August 23--28, 2020, Proceedings, Part X 16}, 41--58. Springer.

\bibitem[{Zamir et~al.(2022)Zamir, Arora, Khan, Hayat, Khan, and Yang}]{restormer}
Zamir, S.~W.; Arora, A.; Khan, S.; Hayat, M.; Khan, F.~S.; and Yang, M.-H. 2022.
\newblock Restormer: Efficient transformer for high-resolution image restoration.
\newblock In \emph{Proceedings of the IEEE/CVF conference on computer vision and pattern recognition}, 5728--5739.

\bibitem[{Zamir et~al.(2020{\natexlab{a}})Zamir, Arora, Khan, Hayat, Khan, Yang, and Shao}]{cycleisp}
Zamir, S.~W.; Arora, A.; Khan, S.; Hayat, M.; Khan, F.~S.; Yang, M.-H.; and Shao, L. 2020{\natexlab{a}}.
\newblock Cycleisp: Real image restoration via improved data synthesis.
\newblock In \emph{Proceedings of the IEEE/CVF conference on computer vision and pattern recognition}, 2696--2705.

\bibitem[{Zamir et~al.(2020{\natexlab{b}})Zamir, Arora, Khan, Hayat, Khan, Yang, and Shao}]{MIRNet}
Zamir, S.~W.; Arora, A.; Khan, S.; Hayat, M.; Khan, F.~S.; Yang, M.-H.; and Shao, L. 2020{\natexlab{b}}.
\newblock Learning enriched features for real image restoration and enhancement.
\newblock In \emph{Computer Vision--ECCV 2020: 16th European Conference, Glasgow, UK, August 23--28, 2020, Proceedings, Part XXV 16}, 492--511. Springer.

\bibitem[{Zamir et~al.(2021)Zamir, Arora, Khan, Hayat, Khan, Yang, and Shao}]{MPRNet}
Zamir, S.~W.; Arora, A.; Khan, S.; Hayat, M.; Khan, F.~S.; Yang, M.-H.; and Shao, L. 2021.
\newblock Multi-stage progressive image restoration.
\newblock In \emph{Proceedings of the IEEE/CVF conference on computer vision and pattern recognition}, 14821--14831.

\bibitem[{Zhang et~al.(2019)Zhang, Dai, Li, and Koniusz}]{DMPHN}
Zhang, H.; Dai, Y.; Li, H.; and Koniusz, P. 2019.
\newblock Deep stacked hierarchical multi-patch network for image deblurring.
\newblock In \emph{Proceedings of the IEEE/CVF conference on computer vision and pattern recognition}, 5978--5986.

\bibitem[{Zhang and Patel(2018)}]{Test1200}
Zhang, H.; and Patel, V.~M. 2018.
\newblock Density-aware single image de-raining using a multi-stream dense network.
\newblock In \emph{Proceedings of the IEEE conference on computer vision and pattern recognition}, 695--704.

\bibitem[{Zhang, Sindagi, and Patel(2019)}]{Test100}
Zhang, H.; Sindagi, V.; and Patel, V.~M. 2019.
\newblock Image de-raining using a conditional generative adversarial network.
\newblock \emph{IEEE transactions on circuits and systems for video technology}, 30(11): 3943--3956.

\bibitem[{Zhang et~al.(2022{\natexlab{a}})Zhang, Li, Zuo, Zhang, Van~Gool, and Timofte}]{DePrior}
Zhang, K.; Li, Y.; Zuo, W.; Zhang, L.; Van~Gool, L.; and Timofte, R. 2022{\natexlab{a}}.
\newblock Plug-and-Play Image Restoration With Deep Denoiser Prior.
\newblock \emph{IEEE Transactions on Pattern Analysis and Machine Intelligence}, 44(10): 6360--6376.

\bibitem[{Zhang et~al.(2020)Zhang, Luo, Zhong, Ma, Stenger, Liu, and Li}]{DBGAN}
Zhang, K.; Luo, W.; Zhong, Y.; Ma, L.; Stenger, B.; Liu, W.; and Li, H. 2020.
\newblock Deblurring by realistic blurring.
\newblock In \emph{Proceedings of the IEEE/CVF conference on computer vision and pattern recognition}, 2737--2746.

\bibitem[{Zhang et~al.(2022{\natexlab{b}})Zhang, Zeng, Guo, and Zhang}]{ELAN}
Zhang, X.; Zeng, H.; Guo, S.; and Zhang, L. 2022{\natexlab{b}}.
\newblock Efficient long-range attention network for image super-resolution.
\newblock In \emph{European Conference on Computer Vision}, 649--667. Springer.

\bibitem[{Zou et~al.(2024{\natexlab{a}})Zou, Chen, Zhang, Li, Ma, Liu, Wang, and Zhang}]{zou2024contourlet}
Zou, Y.; Chen, Z.; Zhang, Z.; Li, X.; Ma, L.; Liu, J.; Wang, P.; and Zhang, Y. 2024{\natexlab{a}}.
\newblock Contourlet Refinement Gate Framework for Thermal Spectrum Distribution Regularized Infrared Image Super-Resolution.
\newblock \emph{arXiv preprint arXiv:2411.12530}.

\bibitem[{Zou et~al.(2024{\natexlab{b}})Zou, Li, Jiang, and Liu}]{zou2024enhancing}
Zou, Y.; Li, X.; Jiang, Z.; and Liu, J. 2024{\natexlab{b}}.
\newblock Enhancing Neural Radiance Fields with Adaptive Multi-Exposure Fusion: A Bilevel Optimization Approach for Novel View Synthesis.
\newblock In \emph{Proceedings of the AAAI Conference on Artificial Intelligence}, volume~38, 7882--7890.

\bibitem[{Zuo, Zhang, and Zhang(2018)}]{dncnn}
Zuo, W.; Zhang, K.; and Zhang, L. 2018.
\newblock \emph{Convolutional Neural Networks for Image Denoising and Restoration}, 93--123.
\newblock Cham: Springer International Publishing.
\newblock ISBN 978-3-319-96029-6.

\end{thebibliography}

\end{document}